\documentclass{article}

\usepackage{microtype}
\usepackage{graphicx}
\usepackage{subcaption}
\usepackage{booktabs} % for professional tables
\usepackage{float}
\usepackage{xfrac}

\usepackage{hyperref}

\usepackage[accepted]{icml2026}

\usepackage{amsmath}
\usepackage{amssymb}
\usepackage{mathtools}
\usepackage{amsthm}

\usepackage{tikz}
\usetikzlibrary{arrows.meta,positioning,fit,calc,shapes.geometric}

\usepackage[capitalize,noabbrev]{cleveref}

\theoremstyle{plain}

\theoremstyle{definition}

\theoremstyle{remark}

\newcommand{\enc}{\text{enc}}
\newcommand{\dec}{\text{dec}}
\newcommand{\aux}{\text{aux}}
\newcommand{\argmax}{\text{argmax}}

\usepackage[textsize=tiny]{todonotes}

\icmltitlerunning{SynthSAEBench: Evaluating Sparse Autoencoders on Scalable Realistic Synthetic Data}

\begin{document}

\twocolumn[
  \icmltitle{SynthSAEBench: Evaluating Sparse Autoencoders \\ on Scalable Realistic Synthetic Data}

  % It is OKAY to include author information, even for blind submissions: the
  % style file will automatically remove it for you unless you've provided
  % the [accepted] option to the icml2026 package.

  % List of affiliations: The first argument should be a (short) identifier you
  % will use later to specify author affiliations Academic affiliations
  % should list Department, University, City, Region, Country Industry
  % affiliations should list Company, City, Region, Country

  % You can specify symbols, otherwise they are numbered in order. Ideally, you
  % should not use this facility. Affiliations will be numbered in order of
  % appearance and this is the preferred way.
  \icmlsetsymbol{equal}{*}

  \begin{icmlauthorlist}
    \icmlauthor{David Chanin}{ucl,mats,decode}
    \icmlauthor{Adrià Garriga-Alonso}{mats}
  \end{icmlauthorlist}

  \icmlaffiliation{ucl}{University College London}
  \icmlaffiliation{mats}{MATS}
  \icmlaffiliation{decode}{Decode Research}

  \icmlcorrespondingauthor{David Chanin}{david.chanin.22@ucl.ac.uk}

  % You may provide any keywords that you find helpful for describing your
  % paper; these are used to populate the "keywords" metadata in the PDF but
  % will not be shown in the document
  \icmlkeywords{Interpretability, Sparse Autoencoders}

  \vskip 0.3in
]

% this must go after the closing bracket ] following \twocolumn[ ...

% This command actually creates the footnote in the first column listing the
% affiliations and the copyright notice. The command takes one argument, which
% is text to display at the start of the footnote. The \icmlEqualContribution
% command is standard text for equal contribution. Remove it (just {}) if you
% do not need this facility.

% Use ONE of the following lines. DO NOT remove the command.
% If you have no special notice, KEEP empty braces:
\printAffiliationsAndNotice{}  % no special notice (required even if empty)
% Or, if applicable, use the standard equal contribution text:
% \printAffiliationsAndNotice{\icmlEqualContribution}

\begin{abstract}
Improving Sparse Autoencoders (SAEs) requires benchmarks that can precisely validate architectural innovations. Current LLM-based SAE benchmarks are too noisy to differentiate architectural improvements, while commonly used synthetic-data experiments are too small-scale, unstandardized, and unrealistic to be meaningful. We introduce \textbf{SynthSAEBench}, a benchmark and toolkit for evaluating SAEs against large-scale synthetic data with realistic feature characteristics including correlation, hierarchy, and superposition, while providing ground-truth features and firings. SynthSAEBench acts as a controlled lower-bound test: SAE architectures that fail when the Linear Representation Hypothesis holds by construction have little hope on real LLMs. The benchmark reproduces known LLM SAE phenomena including the disconnect between reconstruction and latent quality, poor SAE probing, and a precision--recall trade-off mediated by L0, demonstrating that SynthSAEBench findings reproduce results on LLM SAEs. We further identify a novel failure mode: Matching Pursuit SAEs exploit superposition noise to improve reconstruction without learning ground-truth features, suggesting more expressive encoding procedures can easily overfit. SynthSAEBench complements LLM benchmarks with ground-truth features and controlled ablations for diagnosing SAE failure modes, while providing a clear target for SAE architecture work.
\end{abstract}

\section{Introduction}

Large language models (LLMs) achieve remarkable performance but remain opaque, motivating interpretability research into how these models represent knowledge. The Linear Representation Hypothesis (LRH) \citep{park2024lrh} posits that concepts (hereafter ``features'') are represented as nearly-orthogonal linear directions. Models can represent many more features than dimensions by allowing non-orthogonal directions, a phenomenon known as superposition~\citep{elhage2022toy}. Superposition is efficient but makes interpreting activations difficult, motivating the use of Sparse Autoencoders (SAEs) \citep{bricken2023towards,huben2024sparse} to recover underlying feature directions via sparse dictionary learning.

\begin{figure}[!t]
\centering
\resizebox{\columnwidth}{!}{%
\begin{tikzpicture}[
    >=stealth,
    treenode/.style={
        circle,
        draw,
        minimum size=20pt,
        inner sep=1pt,
        font=\tiny
    },
    treeedge/.style={
        draw,
        thick
    },
    sectionlabel/.style={
        font=\normalsize,
        anchor=south
    }
]

% Title
\node[font=\large] at (0, 4.5) {SynthSAEBench feature characteristics};

% ============ HIERARCHY SECTION ============
\node[sectionlabel] at (0, 3.1) {Hierarchy};

% Tree nodes
\node[treenode] (animal) at (0, 2.6) {Animal};
\node[treenode] (dog) at (-1.4, 1.6) {Dog};
\node[treenode] (bird) at (1.4, 1.6) {Bird};
\node[treenode] (poodle) at (-2.1, 0.5) {Poodle};
\node[treenode] (husky) at (-0.7, 0.5) {Husky};
\node[treenode] (eagle) at (0.7, 0.5) {Eagle};
\node[treenode] (hawk) at (2.1, 0.5) {Hawk};

% Tree edges
\draw[treeedge] (animal) -- (dog);
\draw[treeedge] (animal) -- (bird);
\draw[treeedge] (dog) -- (poodle);
\draw[treeedge] (dog) -- (husky);
\draw[treeedge] (bird) -- (eagle);
\draw[treeedge] (bird) -- (hawk);

% ============ CORRELATION SECTION ============
% Label closer to the matrix
\node[sectionlabel] at (-1.7, -0.9) {Correlation};

% Correlation matrix - 5x5 grid
\def\cellsize{0.55}
\def\matrixX{-3.1}
\def\matrixY{-3.7}

% Define colors
\definecolor{darkblue}{RGB}{0,0,139}
\definecolor{medblue}{RGB}{100,100,200}
\definecolor{lightblue}{RGB}{200,200,255}
\definecolor{vlightblue}{RGB}{230,230,255}
\definecolor{vlightpink}{RGB}{255,245,245}
\definecolor{lightpink}{RGB}{255,220,220}
\definecolor{medpink}{RGB}{255,180,180}

% Draw filled squares (row 0 = top row, row 4 = bottom row)
% Row 0 (top)
\fill[darkblue] (\matrixX, \matrixY+4*\cellsize) rectangle +(\cellsize, \cellsize);
\fill[vlightblue] (\matrixX+\cellsize, \matrixY+4*\cellsize) rectangle +(\cellsize, \cellsize);
\fill[lightpink] (\matrixX+2*\cellsize, \matrixY+4*\cellsize) rectangle +(\cellsize, \cellsize);
\fill[vlightblue] (\matrixX+3*\cellsize, \matrixY+4*\cellsize) rectangle +(\cellsize, \cellsize);
\fill[medpink] (\matrixX+4*\cellsize, \matrixY+4*\cellsize) rectangle +(\cellsize, \cellsize);

% Row 1
\fill[vlightblue] (\matrixX, \matrixY+3*\cellsize) rectangle +(\cellsize, \cellsize);
\fill[darkblue] (\matrixX+\cellsize, \matrixY+3*\cellsize) rectangle +(\cellsize, \cellsize);
\fill[vlightblue] (\matrixX+2*\cellsize, \matrixY+3*\cellsize) rectangle +(\cellsize, \cellsize);
\fill[vlightpink] (\matrixX+3*\cellsize, \matrixY+3*\cellsize) rectangle +(\cellsize, \cellsize);
\fill[vlightblue] (\matrixX+4*\cellsize, \matrixY+3*\cellsize) rectangle +(\cellsize, \cellsize);

% Row 2
\fill[lightpink] (\matrixX, \matrixY+2*\cellsize) rectangle +(\cellsize, \cellsize);
\fill[vlightblue] (\matrixX+\cellsize, \matrixY+2*\cellsize) rectangle +(\cellsize, \cellsize);
\fill[darkblue] (\matrixX+2*\cellsize, \matrixY+2*\cellsize) rectangle +(\cellsize, \cellsize);
\fill[vlightblue] (\matrixX+3*\cellsize, \matrixY+2*\cellsize) rectangle +(\cellsize, \cellsize);
\fill[lightpink] (\matrixX+4*\cellsize, \matrixY+2*\cellsize) rectangle +(\cellsize, \cellsize);

% Row 3
\fill[vlightblue] (\matrixX, \matrixY+\cellsize) rectangle +(\cellsize, \cellsize);
\fill[vlightpink] (\matrixX+\cellsize, \matrixY+\cellsize) rectangle +(\cellsize, \cellsize);
\fill[vlightblue] (\matrixX+2*\cellsize, \matrixY+\cellsize) rectangle +(\cellsize, \cellsize);
\fill[darkblue] (\matrixX+3*\cellsize, \matrixY+\cellsize) rectangle +(\cellsize, \cellsize);
\fill[vlightpink] (\matrixX+4*\cellsize, \matrixY+\cellsize) rectangle +(\cellsize, \cellsize);

% Row 4 (bottom)
\fill[medpink] (\matrixX, \matrixY) rectangle +(\cellsize, \cellsize);
\fill[vlightblue] (\matrixX+\cellsize, \matrixY) rectangle +(\cellsize, \cellsize);
\fill[lightpink] (\matrixX+2*\cellsize, \matrixY) rectangle +(\cellsize, \cellsize);
\fill[vlightpink] (\matrixX+3*\cellsize, \matrixY) rectangle +(\cellsize, \cellsize);
\fill[darkblue] (\matrixX+4*\cellsize, \matrixY) rectangle +(\cellsize, \cellsize);

% Draw outer border
\draw[thin] (\matrixX, \matrixY) rectangle (\matrixX+5*\cellsize, \matrixY+5*\cellsize);

% ============ SUPERPOSITION SECTION ============
% Label closer to the box
\node[sectionlabel] at (1.7, -0.9) {Superposition};

% Superposition box
\def\superX{0.35}
\def\superY{-3.7}
\def\superSize{2.75}

% Draw box
\draw[thin] (\superX, \superY) rectangle (\superX+\superSize, \superY+\superSize);

% Draw dashed axes
\draw[dashed, gray!60] (\superX+\superSize/2, \superY) -- (\superX+\superSize/2, \superY+\superSize);
\draw[dashed, gray!60] (\superX, \superY+\superSize/2) -- (\superX+\superSize, \superY+\superSize/2);

% Center point
\pgfmathsetmacro{\cx}{\superX+\superSize/2}
\pgfmathsetmacro{\cy}{\superY+\superSize/2}
\def\radius{1.1}

% Draw 6 vectors radiating from center, equally spaced (60 degrees apart)
\definecolor{vecblue}{RGB}{30,100,200}

\foreach \angle in {15, 75, 135, 195, 255, 315} {
    \draw[vecblue, thick] (\cx, \cy) -- ({\cx+\radius*cos(\angle)}, {\cy+\radius*sin(\angle)});
    \fill[vecblue] ({\cx+\radius*cos(\angle)}, {\cy+\radius*sin(\angle)}) circle (3pt);
}

\end{tikzpicture}
}
\caption{SynthSAEBench provides a large-scale synthetic data model with realistic feature characteristics including correlation, hierarchy, superposition and zipfian firing distributions, scalable to hundreds of thousands of features and realistic hidden dimension sizes.}
\label{fig:overview}
\end{figure}
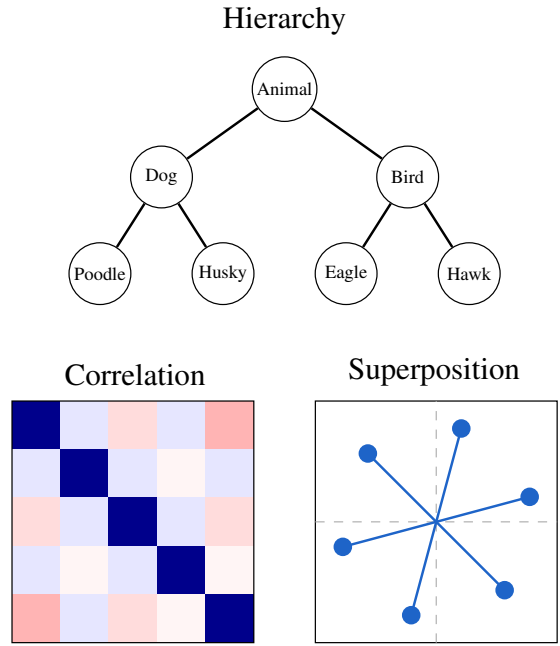

A key challenge in improving SAEs is that we lack ground-truth knowledge of the ``true features'' in an LLM. LLM benchmarks such as SAEBench \citep{karvonen2025saebenchcomprehensivebenchmarksparse} evaluate SAE performance on tasks like sparse probing \citep{gurnee2023finding,kantamneni2025sparse}, concept disentanglement \citep{karvonen2024evaluating}, and autointerpretability \citep{paulo2025automatically}. However, SAEBench metrics exhibit substantial noise between runs (see Appendix~\ref{apx:noise-in-saebench-metrics}), making it difficult to evaluate small architectural improvements. Moreover, without ground-truth access, we cannot diagnose \emph{why} SAEs score poorly, a critical obstacle given recent work showing that SAEs underperform supervised methods like logistic-regression probes~\citep{kantamneni2025sparse}.

On the other extreme, SAE research already relies on bespoke synthetic models: toy experiments with fewer than 10 independent features \citep{song2025position,Gribonval2010Dictionary,elhage2022toy}, or per-paper setups studying hierarchy \citep{chanin2025a,costa2025flat,bussmann2025learning} and correlation \citep{chanin2025feature,chanin2025sparse}. Nearly all SAE architecture work already involves experiments with synthetic models, but these models are too small, unrealistic, and unstandardized to compare architectures.

We introduce \textbf{SynthSAEBench}, an extension of the SAELens library~\citep{bloom2024saetrainingcodebase} that scales the synthetic model approach to 16k+ features with correlation, hierarchy, and superposition, providing ground-truth feature directions and firings. Data generation runs at hundreds of thousands of samples per second on a single GPU. We release a standard pretrained configuration\footnote{\url{https://huggingface.co/decoderesearch/synth-sae-bench-16k-v1}}
 as the canonical benchmark, while allowing practitioners to change any part of the synthetic model to perform ablations and test SAE performance under different feature assumptions.

SynthSAEBench is a controlled lower-bound test for SAE architectures: If SAEs fail when the Linear Representation Hypothesis holds by construction -- the setting SAEs are explicitly designed around -- then we cannot expect them to succeed on real LLMs. Indeed, most previously observed LLM SAE phenomena also emerge in this regime: (1) Matryoshka SAEs overperform on SAEBench despite poor reconstruction~\citep{bussmann2025learning,karvonen2025saebenchcomprehensivebenchmarksparse}, (2) Matching Pursuit SAEs overperform on reconstruction while scoring poorly on SAEBench~\citep{chanin2025trainingmpsaes}, (3) poor SAE probing performance~\citep{kantamneni2025sparse}, (4) a precision-recall trade-off in probing mediated by SAE L0~\citep{chanin2025a}, and (5) BatchTopK and JumpReLU both being nearly equivalent state-of-the-art architectures~\citep{bloom2024saetrainingcodebase,bussmann2024batchtopk,rajamanoharan2024jumping}. This shows that performance on SynthSAEBench transfers to results on real LLM SAEs.

Additionally, using SynthSAEBench, we find that Matching Pursuit SAEs exploit superposition noise to improve reconstruction without learning ground-truth features, hinting at why the simple linear encoder of traditional SAEs is so hard to outperform despite known theoretical limitations~\citep{oneill2025compute}.

No SAE architecture we evaluate achieves perfect performance on SynthSAEBench, highlighting a clear target for improvement in SAE architectures.

% Code is available on Anonymous Github\footnote{Code: \url{https://anonymous.4open.science/r/synth-sae-bench-anon-0C61}}.

\section{Background}

\paragraph{Sparse autoencoders (SAEs).} An SAE decomposes an input activation \(a \in \mathbb{R}^D\) into a hidden state \(f\) consisting of \(L\) hidden neurons, called ``latents''. An SAE is composed of an encoder \(W_\enc \in \mathbb{R}^{L \times D}\), a decoder \(W_\dec \in \mathbb{R}^{D \times L} \), a decoder bias \(b_\dec \in \mathbb{R}^D\), and an encoder bias \(b_\enc \in \mathbb{R}^L\), and a nonlinearity \(\sigma\), typically ReLU or a variant like JumpReLU \citep{rajamanoharan2024jumping}, TopK \citep{gao2024scaling} or BatchTopK (BTK) \citep{bussmann2024batchtopk}.

\begin{align}
f = & \sigma(W_\enc (a - b_\dec) + b_\enc) \\
\hat{a} = & W_\dec f + b_\dec
\end{align}

The SAE is trained with a reconstruction loss, typically Mean Squared Error (MSE), and a sparsity-inducing loss consisting of a function \(\mathcal{S}\) that penalizes non-sparse representation with corresponding sparsity coefficient \(\lambda\). For standard L1 SAEs, \(\mathcal{S}\) is the L1 norm of \(f\). For TopK and BatchTopK SAEs, there is no sparsity-inducing loss (\(\mathcal{S}=0\)) as the TopK function directly induces sparsity. There is sometimes also an additional auxiliary loss \(\mathcal{L}_{aux}\) with coefficient \(\alpha\) to ensure all latents fire. Standard L1 SAEs typically do not have an auxiliary loss \citep{olah2024april}. The general SAE loss is

\begin{equation}
        \mathcal{L} = \|a - \hat{a}\|_2^2 + \lambda \mathcal{S} + \alpha \mathcal{L}_\aux.
\end{equation}

\paragraph{Matryoshka SAEs.} A matryoshka SAE \citep{bussmann2025learning} extends the SAE definition by summing losses created by prefixes of SAE latents. This forces each sub-SAE to reconstruct input activations on its own, and incentivizes the SAE to place more common, general concepts into latents with smaller index number. A matryoshka SAE uses nested prefixes with sizes \(\mathcal{M} = m_1, m_2, ... m_n\) where \(m_1 < m_2 < \ldots < m_n = L\), where \(L\) is the number of latents in the full dictionary. Matryoshka SAE loss is:

\begin{equation}
    \mathcal{L} = \sum_{m \in \mathcal{M}} \left( \|a - \hat{a}_{m}\|_2^2 + \lambda \mathcal{S}_{m} \right) + \alpha \mathcal{L}_\aux
    \label{eq:matryoshka}
\end{equation}

Where \(\hat{a}_m\) is the reconstruction for the SAE using the first \(m\) latents, and \(\mathcal{S}_m\) is the sparsity penalty applied to the first \(m\) latents. For TopK and BatchTopK Matryoshka SAEs, there is no sparsity penalty (\(\mathcal{S}_m=0\)) as the TopK function directly imposes sparsity.

\paragraph{Matching Pursuit SAEs.} A Matching Pursuit (MP) SAE \citep{costa2025flat} acts like a TopK SAE where the $k$ latents are selected in serial rather than in parallel. In an MP-SAE, there is no explicit encoder $W_\enc$ or encoder bias $b_\enc$. Instead, at each iteration $t \in \{0, \dots, k-1\}$, the latent with the highest projection onto the residual $r_t$ is selected and its contribution is projected out:

\begin{align}
    l_t &= \argmax_i\, W_{\dec,i}^\top r_t \\
    r_{t+1} &= r_t - \bigl(W_{\dec,l_t}^\top r_t\bigr)\, W_{\dec,l_t}
\end{align}

with $r_0 = a$. The SAE latent vector $f$ accumulates the projection coefficients across iterations into the corresponding latent indices:
\begin{equation}
    f_l = \sum_{t\,:\,l_t = l} W_{\dec,l_t}^\top r_t,
\end{equation}
so a latent selected on multiple iterations has its activations summed. The MP-SAE is trained with the standard MSE reconstruction loss $\|a - \hat{a}\|_2^2$, where the MP-SAE reconstruction is $\hat{a} = W_\dec f$ (no decoder bias); by construction, this equals $\|r_k\|_2^2$.

The variant of MP-SAEs we use in this paper does not do any early stopping based on $\|r\|_2$ or based on selecting the same latent multiple times, as early stopping adds more complication to the training process and has not been shown to improve results \citep{chanin2025trainingmpsaes}.

\begin{figure*}[t]
\centering
\resizebox{\textwidth}{!}{%
\input{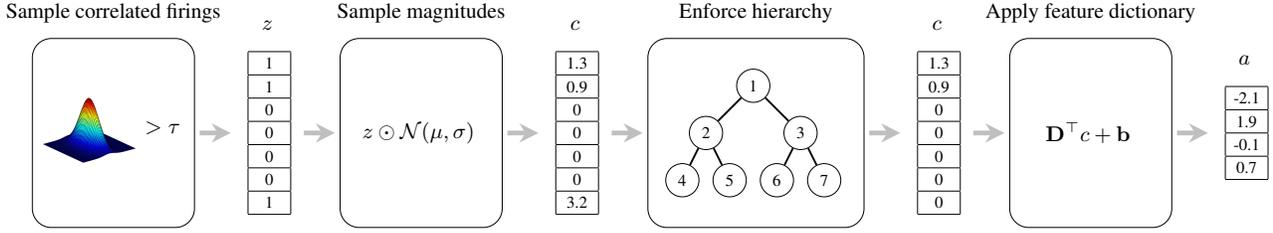}
}
\caption{Overview of process to generate a single training activation, $a$.}
\label{fig:sampling}
\end{figure*}

\section{Synthetic Data Model}

We now describe our synthetic data model in detail. We extend the traditional bernoulli-gaussian model commonly used in dictionary learning literature~\citep{Gribonval2010Dictionary,wang2020unique} and follows the LRH by construction. This is also the standard synthetic setting used by SAE research~\citep{song2025position,oneill2025compute,bussmann2025learning}. 

Our model contains a feature dictionary $\mathbf{D} \in \mathbb{R}^{N \times D}$ containing $N$ ground-truth feature directions, each represented as a unit-norm vector $\mathbf{d}_i \in \mathbb{R}^D$, an optional bias $\mathbf{b} \in \mathbb{R}^D$, and an activation generator that samples sparse feature coefficients. Each feature $d_i$ has a corresponding firing probability $p_i$.

Given a sample, the activation generator first determines which features fire using a Gaussian copula approach to generate correlated binary firing indicators. First, we sample from a multivariate Gaussian with correlation structure $\boldsymbol{\Sigma}$:
\begin{equation}
\mathbf{g} \sim \mathcal{N}(\mathbf{0}, \boldsymbol{\Sigma}).
\end{equation}
We then threshold these samples to obtain binary firing indicators that respect both the marginal firing probabilities $p_i$ and the correlation structure. For feature $i$ with firing probability $p_i$, we compute the threshold as $\tau_i = \Phi^{-1}(1 - p_i)$, where $\Phi^{-1}$ is the inverse standard normal CDF, and set $z_i = \mathbf{1}[g_i > \tau_i]$. When $\boldsymbol{\Sigma} = \mathbf{I}$ (no correlations between features), this is equivalent to $z_i \sim \text{Bernoulli}(p_i)$.

For features that fire ($z_i = 1$), coefficients are sampled from a rectified Gaussian distribution:
\begin{align}
c_i &= z_i \cdot \text{ReLU}(\mu_i + \sigma_i \epsilon_i), \quad \epsilon_i \sim \mathcal{N}(0, 1),
\end{align}
where $\mu_i$ and $\sigma_i$ are the per-feature mean and standard deviation of firing magnitudes. Optionally, a post-processing function $h: \mathbb{R}^N \to \mathbb{R}^N$ can be applied to modify the coefficient vector, i.e., $\mathbf{c} \leftarrow h(\mathbf{c})$; we use this mechanism to implement hierarchical constraints (\S\ref{sec:hierarchy}). 

The hidden activation is then computed as:
\begin{equation}
\mathbf{a} = \sum_{i=1}^{N} c_i \mathbf{d}_i + \mathbf{b} = \mathbf{D}^\top \mathbf{c} + \mathbf{b}.
\end{equation}

For scalability with large numbers of features, we use a low-rank correlation matrix:
\begin{equation}
\boldsymbol{\Sigma} = \mathbf{F}\mathbf{F}^\top + \text{diag}(\boldsymbol{\delta}),
\end{equation}
where $\mathbf{F} \in \mathbb{R}^{N \times r}$ is a factor matrix of rank $r \ll N$, and $\boldsymbol{\delta} \in \mathbb{R}^N$ contains diagonal residual variances chosen to ensure unit diagonal in $\boldsymbol{\Sigma}$ (i.e., $\delta_i = 1 - \sum_j F_{ij}^2$). Sampling from this structure is efficient:
\begin{equation}
\mathbf{g} = \mathbf{F}\boldsymbol{\epsilon} + \sqrt{\boldsymbol{\delta}} \odot \boldsymbol{\eta}, \quad \boldsymbol{\epsilon} \sim \mathcal{N}(\mathbf{0}, \mathbf{I}_r), \quad \boldsymbol{\eta} \sim \mathcal{N}(\mathbf{0}, \mathbf{I}_N),
\end{equation}
where $\odot$ denotes elementwise multiplication. This requires only $O(Nr)$ computation per sample rather than $O(N^2)$ for full covariance sampling.

This generative model enables us to control several important phenomena that arise in real neural networks: superposition (\S\ref{sec:superposition}), feature correlation (\S\ref{sec:correlation}), and feature hierarchy (\S\ref{sec:hierarchy}). Both per-feature firing probabilities $p_i$ and magnitude parameters $(\mu_i, \sigma_i)$ admit configurable distributions; we describe the supported options in Appendix~\ref{apx:firing-dists}. Figure~\ref{fig:sampling} shows the data generating process.

% \begin{figure}[t]
% \centering
% \resizebox{\columnwidth}{!}{%
% \input{figures/graphical_model.tex}
% }
% \caption{Bayesian graphical model of the synthetic data generating process. Rectangles denote parameters, circles denote random variables (shaded = observed, double = deterministic function of parents). The inner plate iterates over $N$ features; the outer plate iterates over $B$ samples. For each sample: correlated Gaussian samples $g_i \sim \mathcal{N}(0, \boldsymbol{\Sigma})$ are thresholded using firing probabilities $p_i$ to obtain binary indicators $z_i$; magnitude noise $\epsilon_i \sim \mathcal{N}(0,1)$ combines with $z_i$ and per-feature means $\mu_i$, standard deviations $\sigma_i$ to produce coefficients $c_i$; finally, the activation $\mathbf{a}$ is computed as a linear combination using feature dictionary $\mathbf{D}$ and bias $\mathbf{b}$.}
% \label{fig:graphical_model}
% \end{figure}

\subsection{Superposition}
\label{sec:superposition}

The Linear Representation Hypothesis~\citep{park2024lrh} posits that neural networks represent more concepts than they have dimensions, forcing features to share representational capacity---a phenomenon known as superposition~\citep{elhage2022toy}. We characterize the degree of superposition by \emph{mean max absolute cosine similarity}, $\rho_{\text{mm}}$. $\rho_{\text{mm}} = \frac{1}{N} \sum_{i=1}^{N} \max_{j \neq i} |\mathbf{d}_i^\top \mathbf{d}_j|$. $0 \le \rho_{\text{mm}} \le 1$, with $\rho_{\text{mm}} = 0$ indicating no superposition (i.e., all features are mutually orthogonal).

Feature vectors are initialized as random unit vectors sampled from a standard normal distribution:
\begin{equation}
\mathbf{d}_i \leftarrow \frac{\mathbf{g}_i}{\|\mathbf{g}_i\|_2}, \quad \mathbf{g}_i \sim \mathcal{N}(\mathbf{0}, \mathbf{I}_D).
\end{equation}

While this initialization produces features with small expected pairwise cosine similarities (scaling as $O(1/\sqrt{D})$), some feature pairs may have higher overlap by chance. To reduce spurious correlations between feature directions, we optionally apply an orthogonalization procedure to minimize pairwise cosine similarity. Specifically, we optimize:
\begin{equation}
\mathcal{L}_\text{ortho} = \sum_{i \neq j} (\mathbf{d}_i^\top \mathbf{d}_j)^2 + \lambda \sum_i (\|\mathbf{d}_i\|_2 - 1)^2
\end{equation}
using gradient descent. After orthogonalization, all vectors are renormalized to unit length. This procedure encourages feature vectors to be as orthogonal as possible given the dimensionality constraints, although when using models with a large hidden dimension, we find that random initialization already results in features that are nearly orthogonal.

For scalability with large $N$, we use a memory-efficient chunked implementation that computes pairwise dot products in blocks. This reduces memory complexity from $O(N^2)$ to $O(\text{chunk\_size} \times N)$, enabling orthogonalization of dictionaries with thousands or even millions of features.

\subsection{Feature Correlation}
\label{sec:correlation}

Real neural network features are rarely independent: concepts that co-occur in data tend to co-fire in the network's representations. Previous work has shown that correlated features present challenges for SAEs, leading to phenomena like feature hedging \citep{chanin2025feature,chanin2025sparse}. To study these effects systematically, we implement configurable correlation structures between feature firings.

We support randomly generating a low-rank correlation matrix for use in the synthetic model. This random generation is controlled by two parameters: the rank $r$ (which determines the complexity of the correlation patterns) and a correlation scale $s$ (which controls the magnitude of off-diagonal correlations by scaling the factor matrix). The detailed generation procedure is described in Appendix~\ref{apx:correlation}.

\subsection{Feature Hierarchy}
\label{sec:hierarchy}

Concepts in natural language and vision exhibit hierarchical structure: ``golden retriever'' is a type of ``dog,'' which is a type of ``animal.'' Previous work has shown that SAEs struggle with hierarchical features, leading to phenomena like feature absorption \citep{chanin2025a}. We implement configurable hierarchical dependencies between features to study these effects.

Our hierarchy is represented as a forest of trees, where each node corresponds to a feature and children can only fire when their parent is active. Formally, after sampling the initial firing indicators $z_i$, we apply the constraint:
\begin{equation}
c_{\text{child}} \leftarrow c_{\text{child}} \cdot \mathbf{1}[c_{\text{parent}} > 0],
\end{equation}
which zeros out child activations whenever the parent is inactive.

Additionally, we support \emph{mutual exclusion} among siblings: when a parent node is marked as having mutually exclusive children, at most one child can be active per sample. This models concepts like ``dog'' vs. ``bird''---both are animals, but a single entity cannot be both simultaneously. When multiple siblings would fire, one is randomly selected as the winner and the others are deactivated.

We also support \emph{parent-scaled magnitudes}, where child activation magnitudes are modulated by their parent's activation strength rather than simply being binary-gated. This models the intuition that the intensity of a specific concept (e.g., ``golden retriever'') should scale with the intensity of its parent concept (e.g., ``dog''). Details are in Appendix~\ref{apx:parent-scaling}.

% The hierarchy is configured with the following parameters:
% \begin{itemize}
% \item \textbf{Root nodes}: Number of independent hierarchy trees.
% \item \textbf{Branching factor}: Number of children per parent node (can be a fixed value or a range).
% \item \textbf{Maximum depth}: How many levels deep the trees extend.
% \item \textbf{Mutual exclusion portion}: Fraction of parent nodes whose children are mutually exclusive.
% \item \textbf{Mutual exclusion depth range}: Which depth levels can have mutual exclusion.
% \end{itemize}

For efficiency, we precompute the hierarchy structure as sparse index tensors, enabling $O(\text{active features})$ processing rather than $O(N)$ per sample.

\section{Evaluating SAEs on Synthetic Data}

A key advantage of synthetic data is access to ground-truth feature directions and activations, enabling precise evaluation of SAE quality. We implement a comprehensive set of metrics organized into four categories: reconstruction quality, feature recovery, classification accuracy, and sparsity.

\subsection{Reconstruction Metrics}

\paragraph{Explained Variance ($R^2$).} We measure how well the SAE reconstruction $\hat{\mathbf{a}}$ captures the variance in the input activations $\mathbf{a}$:
\begin{equation}
R^2 = 1 - \frac{\mathbb{E}[\|\mathbf{a} - \hat{\mathbf{a}}\|_2^2]}{\text{Var}(\mathbf{a})},
\end{equation}
where $\text{Var}(\mathbf{a}) = \mathbb{E}[\|\mathbf{a}\|_2^2] - \|\mathbb{E}[\mathbf{a}]\|_2^2$ is the total variance. A value of 1.0 is perfect reconstruction.

% \paragraph{Shrinkage.} SAEs often exhibit shrinkage, where reconstructions have smaller magnitude than inputs. We measure this as:
% \begin{equation}
% \text{Shrinkage} = \mathbb{E}\left[\frac{\|\hat{\mathbf{a}}\|_2}{\|\mathbf{a}\|_2}\right].
% \end{equation}
% A value of 1.0 indicates no shrinkage.

\subsection{Feature Recovery Metrics}

\paragraph{Mean Correlation Coefficient (MCC).} Following \citet{song2025position}, we measure how well SAE decoder columns $\mathbf{w}_j$ align with ground-truth feature directions $\mathbf{d}_i$ using optimal bipartite matching. We compute the absolute cosine similarity matrix $|S_{ij}| = |\mathbf{w}_j^\top \mathbf{d}_i|$ (both $\mathbf{w}_j$ and $\mathbf{d}_i$ are unit-norm), find the optimal one-to-one matching via the Hungarian algorithm, and report the mean similarity of matched pairs:
\begin{equation}
\text{MCC} = \frac{1}{\min(L, N)} \sum_{(i,j) \in \text{matching}} |\mathbf{w}_j^\top \mathbf{d}_i|.
\end{equation}

\paragraph{Feature Uniqueness.} We measure what fraction of SAE latents track unique ground-truth features. For each latent $j$, we find its best-matching ground-truth feature: $i^*(j) = \arg\max_i |\mathbf{w}_j^\top \mathbf{d}_i|$. Uniqueness is the fraction of unique best matches divided by the number of latents:
\begin{equation}
\text{Uniqueness} = \frac{|\{i^*(j) : j = 1, \ldots, L\}|}{L}.
\end{equation}
A value of 1.0 means each latent tracks a different feature.

\subsection{Classification Metrics}

We evaluate each SAE latent as a binary classifier for its best-matching ground-truth feature. For latent $j$ matched to feature $i^*(j)$, we compute standard classification metrics over evaluation samples: precision, recall, and F1 score. A latent is considered to fire when $f_j > 0$, and a ground-truth feature is considered to fire when $z_i = 1$. We report the mean of each metric across all latents.

\subsection{Sparsity Metrics}

\paragraph{L0 Comparison.} We track the average L0 (number of active features per sample) of both ground-truth activations and for SAE latent activations.

\paragraph{Dead Latents.} We count the number of SAE latents that never fire across the evaluation set. Dead latents represent wasted capacity and indicate training issues.

\section{Benchmark Configuration}

We define the standard SynthSAEBench configuration, chosen to be large-scale enough to elicit realistic SAE training dynamics while remaining tractable to run.

% \begin{figure}
%     \centering
%     \includegraphics[width=0.49\columnwidth]{assets/stats/firing_frequency_histogram.pdf}\hfill
%     \includegraphics[width=0.49\columnwidth]{assets/stats/hierarchy_stats.pdf}
%     \caption{SynthSAEBench feature firing probabilities (left) and hierarchy distribution (right).}
%     \label{fig:dataset-stats}
% \end{figure}

\begin{figure}
    \centering
    \includegraphics[width=\columnwidth]{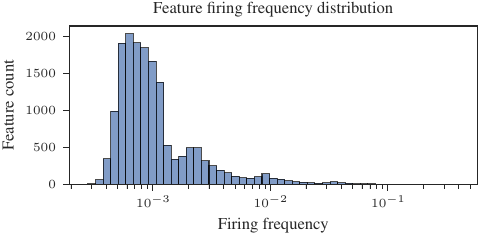}
    \caption{SynthSAEBench-16k feature firing probabilities.}
    \label{fig:firing-probs}
\end{figure}

\begin{figure}
    \centering
    \includegraphics[width=\columnwidth]{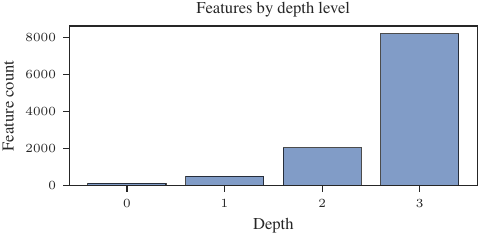}
    \caption{SynthSAEBench-16k hierarchy distribution.}
    \label{fig:hierarchy}
\end{figure}

The model has $N = 16{,}384$ features in hidden dimension $D = 768$, yielding $\rho_\text{mm} \approx 0.15$. We estimate this is similar to the level of superposition 1B features would cause in a Gemma-2-2b sized model (Appendix~\ref{apx:exploring-superposition}). Activations have mean L2 norm $28$ and stdev $5$, matching Pythia-160m~\citep{biderman2023pythia} layer 10. Feature vectors are random unit vectors orthogonalized for 100 steps (lr $3\times10^{-4}$); the bias is a random Gaussian direction normalized to magnitude $0.5$. Base firing probabilities follow a Zipfian distribution ($\alpha=0.5$, $p_{\max}=0.4$, $p_{\min}=5\times10^{-4}$); firing magnitudes have means interpolating linearly from $5.0$ to $4.0$, with stdevs $\sigma_i \sim |\mathcal{N}(0.5, 0.5^2)|$. The firing probability distribution is shown in Figure~\ref{fig:firing-probs}.

The hierarchy has 128 root trees with branching factor 4 and max depth 3, covering 10{,}880 features. All sibling nodes are mutually exclusive, child magnitudes scale with parent activation (Appendix~\ref{apx:parent-scaling}), and base probabilities are compensated for hierarchy and mutual exclusion effects (Appendix~\ref{apx:compensate}). The distribution of nodes in the hierarchy is shown in Figure~\ref{fig:hierarchy}. Beyond the hierarchy, a random low-rank correlation with rank $r=25$ and scale $s=0.1$ adds structured dependencies. The model has an L0 of $34$, determined by the firing-probability distribution.

\subsection{Benchmark instructions}
\label{sec:instructions}

\textit{We recommend training SAEs of width 4096 on SynthSAEBench}, intentionally more narrow than the model dictionary, since LLM SAEs are always narrower than the (unknown) number of true features in an LLM~\citep{templeton2024scaling}. Unless otherwise specified, we train SAEs on 200M samples (15--20 minutes per SAE on a single H100, batch size 1024). Sampling performance is in Appendix~\ref{apx:perf}.

\section{Results}

\begin{figure*}[ht]
  \centering
  \includegraphics[width=\textwidth]{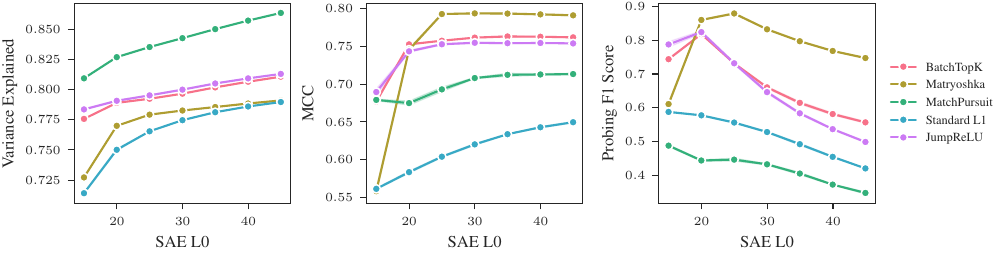}
  \caption{Variance explained (left), MCC (middle), and F1-score (right) for SynthSAEBench SAEs with varying L0. Shaded area is stdev with 5 seeds (too small to be visible for most SAEs).}
  \label{fig:vary-l0}
\end{figure*}

We train standard L1, Matching Pursuit, BatchTopK, Matryoshka BatchTopK, and JumpReLU SAEs on SynthSAEBench.\footnote{Experiment code is available at \url{https://github.com/decoderesearch/synth-sae-bench-experiments}.}
These SAEs have width 4096 and are trained on 200M samples, as recommended in Section~\ref{sec:instructions}. For standard L1 SAEs and JumpReLU SAEs, we implement a controller to automatically tune their sparsity coefficient during training to hit a target L0. This controller is described in more detail in Appendix~\ref{apx:autotune}.

We vary the L0 of the SAEs from 15 to 45 with 5 seeds per L0. F1-score, MCC, and variance explained are shown in Figure~\ref{fig:vary-l0}. Matryoshka SAEs achieve the best probing and MCC scores, indicating the best latent quality, despite poor explained variance (poor reconstruction). This matches the results of Matryoshka SAEs on SAEBench~\citep{karvonen2025saebenchcomprehensivebenchmarksparse}, where this same pattern is observed for LLM SAEs. The performance drop-off for Matryoshka SAEs at low L0 seems to be due to dead latents rather than a fundamental architectural issue (see Appendix~\ref{apx:dead-latents}). We introduce a per-prefix Matryoshka auxiliary loss that reduces, but does not eliminate, these dead latents (Appendix~\ref{apx:matryoshka}). MP-SAEs, on the other hand, have the best variance explained, but have poor probing and MCC results, indicating that MP-SAEs' impressive reconstruction comes at the cost of latent quality. We see JumpReLU and BatchTopK achieving roughly similar levels of quality and reconstruction, further validating why these are both considered state-of-the-art architectures.

\begin{figure}[ht]
  \centering
  \includegraphics[width=\columnwidth]{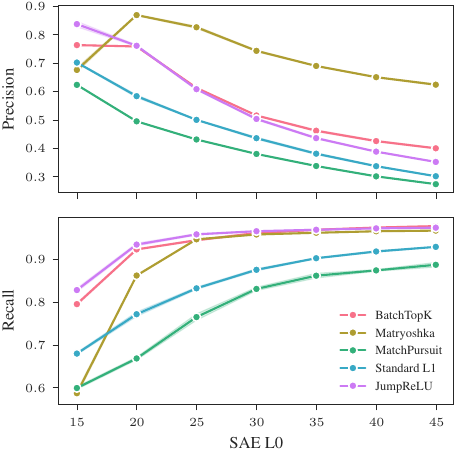}
  \caption{Probing precision and recall for SAEs trained on SynthSAEBench across varying L0 values. Higher L0 increases recall at the cost of precision. Shaded area is stdev.}
  \label{fig:prec-recall}
\end{figure}

\subsection{Precision-recall trade-off mediated by L0}
\label{sec:prec-recall}

One striking result from Figure~\ref{fig:vary-l0} is that no SAE achieves near-perfect probing F1 at any L0, with the best performer, Matryoshka SAEs, peaking at around F1=0.88. This means no SAEs will act as a great classifier for ground-truth model features. This is consistent with results showing that LLM SAEs underperform logistic-regression probes~\citep{kantamneni2025sparse}. Indeed, logistic regression probes trained directly on SynthSAEBench activations achieve a mean F1 of 0.974 (Table~\ref{tab:probe-results}; training details in Appendix~\ref{apx:probes}), substantially outperforming the best SAE.

\begin{table}[ht]
\centering
\caption{Logistic regression probe results on SynthSAEBench (first 4,096 highest-frequency features). Supervised probes substantially outperform the best SAE probing F1 of $\approx$0.88.}
\label{tab:probe-results}
\begin{tabular}{lcc}
\toprule
Metric & Mean & Median \\
\midrule
AUC       & 0.9999 & 1.0000 \\
Accuracy  & 0.9998 & 0.9999 \\
Precision & 0.9800 & 0.9853 \\
Recall    & 0.9680 & 0.9711 \\
F1        & 0.9739 & 0.9775 \\
\bottomrule
\end{tabular}
\end{table} This poor probing performance is one of the key criticisms of SAEs, so our ability to reproduce it in a setting with known ground-truth features is a key first step to address this problem with SAE architectural improvements.

We show the precision and recall for the probing task in Figure~\ref{fig:prec-recall}. Higher L0 increases recall at the cost of precision, reproducing the precision-recall trade-off mediated by SAE L0 seen in previous LLM SAE studies~\citep{karvonen2025saebenchcomprehensivebenchmarksparse,chanin2025a}.

\subsection{Near-equivalence between JumpReLU and BatchTopK}

JumpReLU and BatchTopK are both considered state-of-the-art architectures~\citep{bloom2024saetrainingcodebase,rajamanoharan2024jumping,bussmann2024batchtopk}, and this is borne out on SynthSAEBench as well. Among standard architectures, we see these both performing nearly identically on F1 score at L0=20 and L0=25, and very closely on both MCC and reconstruction. This further shows that SynthSAEBench results reflect what we see in LLM SAEs.

\subsection{MP-SAEs overfit superposition noise}

\begin{figure*}[ht]
  \centering
  \includegraphics[width=\textwidth]{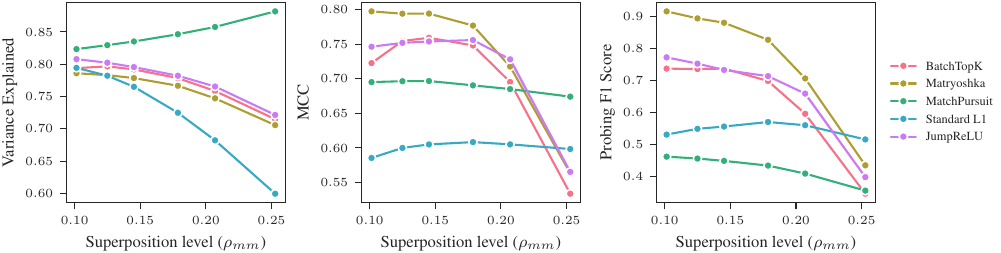}
  \caption{Variance explained (left), MCC (middle), and F1-score (right) for SAEs trained on variants of SynthSAEBench with different levels of superposition. Interestingly, MP-SAEs increase their variance explained at high superposition, implying they are able to overfit on superposition noise.}
  \label{fig:vary-superposition}
\end{figure*}

We next investigate the effect of superposition noise on SAE performance. We modify the hidden dim of the base SynthSAEBench model, ranging from 256 to 1536, scaling training samples by $(\frac{D}{768})^{(0.6)}$ to account for the change in SAE parameters as recommended by~\citet{gao2024scaling}. We train all SAEs at L0=25, as this seems to be a good setting for most SAEs on the base SynthSAEBench model. Results are shown in Figure~\ref{fig:vary-superposition}.

We see the variance explained decrease with increasing superposition noise except, interestingly, for Matching Pursuit (MP) SAEs. MP-SAEs actually increase variance explained the more superposition noise is present while decreasing MCC and F1-score, implying that MP-SAEs are able to overfit feature overlap due to superposition to improve reconstruction. MP-SAEs use a more expressive encoding process than the simple linear encoder of traditional SAEs, and this extra expressivity can result in overfitting reconstruction without learning the underlying features of the model, perhaps explaining why traditional SAEs with simple linear encoders are so hard to outperform in practice~\citep{oneill2025compute}. This insight was only possible because SynthSAEBench provides both ground-truth features (to measure MCC and F1 independently of reconstruction) and controlled superposition levels (to isolate its effect), neither of which is available when benchmarking on LLMs.

\subsection{Additional ablations}
\label{sec:ablations}

We further ablate feature correlation strength and rank, base firing probabilities, firing magnitude stdev, and hierarchy depth (Appendix~\ref{apx:extended}). The overall pattern from Figure~\ref{fig:vary-l0} is stable across all of these ablations: Matryoshka SAEs lead on latent quality (MCC and F1) while MP-SAEs lead on reconstruction. Notably, increasing correlation strength increases variance explained while slightly decreasing probing F1, consistent with prior work showing SAEs exploit correlations by mixing correlated features into a single latent~\citep{chanin2025feature,chanin2025sparse}. Increasing firing magnitude stdev improves MCC while reducing F1 for every SAE: high variance makes feature directions more salient, but low-magnitude firings become hard to disambiguate from superposition noise. Surprisingly, standard L1 SAEs achieve the best MCC of all architectures in this high-stdev regime.

\section{Related Work}

\paragraph{SAE evaluation.} Evaluating SAEs on LLMs is challenging due to the lack of ground truth. SAEBench \citep{karvonen2025saebenchcomprehensivebenchmarksparse} provides a suite of downstream tasks including sparse probing \citep{gurnee2023finding}, concept erasure \citep{karvonen2024evaluating}, and autointerp \citep{paulo2025automatically}. However, these metrics have high variance and measure indirect proxies rather than feature recovery directly. The MCC metric \citep{oneill2025compute} provides a principled way to compare learned features to ground truth. While not directly applicable for SAEs, Interpbench~\cite{gupta2024interpbench} provides a synthetic model for circuit discovery work.

\paragraph{Toy models and synthetic data.} \citet{elhage2022toy} introduced toy models of superposition to study how neural networks represent more features than dimensions. Subsequent work has used small-scale synthetic setups to study specific phenomena: feature absorption in hierarchical features \citep{chanin2025a}, feature hedging under correlation \citep{chanin2025feature}, and incorrect L0 behavior \citep{chanin2025sparse}. However, these studies use bespoke synthetic models that are not standardized or comparable. Our work provides a unified, large-scale synthetic framework that encompasses superposition, correlation, and hierarchy while providing ground-truth features.

\paragraph{Alternative representation hypotheses.} Our work follows the LRH, but other feature hypotheses exist that extend the LRH and are thus natural extensions for our work. \citet{fel2025into} introduce the Minkowski Representation Hypothesis, where the softmax operation in the transformer results in polytopes in the representation space. Another extension is feature manifolds~\citep{chen2018sparse,michaud2025understanding}, where features span manifolds rather than corresponding to a single linear direction. 

\section{Discussion}

SynthSAEBench represents a best-case scenario for SAEs: the Linear Representation Hypothesis holds by construction, with features that are truly linear directions. Yet no SAE architecture we evaluate achieves perfect feature recovery. This is significant because a common response to SAE failures on LLMs is that the representation hypothesis may not hold exactly. Our results show that even when the LRH is perfectly satisfied, current SAE architectures still struggle with superposition, correlation, and hierarchy. The bottleneck is in the SAE architectures and training dynamics themselves, not in the representation hypothesis, strengthening the case for continued architectural innovation.

The SAEs we train are deliberately narrower than the underlying model ($L = 4096 < N = 16{,}384$), reflecting the LLM regime where SAEs are always narrower than the number of true features in the model~\citep{templeton2024scaling}. Under MSE, this capacity mismatch incentivizes learned latents to encode mixtures of ground-truth features~\citep{chanin2025feature,chanin2025sparse}, and this feature mixing or ``hedging'' is precisely a failure mode current SAEs exhibit. We view imperfect recovery on SynthSAEBench not as a confound but as the target problem: the goal is architectures or training procedures that recover features cleanly even when narrower than the underlying model.

With ground-truth features and controlled ablations, we reproduce key LLM SAE phenomena: Matryoshka SAEs' high probing performance despite poor reconstruction~\citep{bussmann2025learning}, MP-SAEs' poor probing despite high reconstruction~\citep{chanin2025trainingmpsaes}, the gap between SAE probing and supervised probes~\citep{kantamneni2025sparse}, and the precision-recall trade-off mediated by L0~\citep{chanin2025a}. We also identify a new failure mode: MP-SAEs exploit superposition noise to improve reconstruction without learning ground-truth features, which may partly explain why traditional SAEs with simple linear encoders remain hard to outperform~\citep{oneill2025compute}; more expressive encoding procedures can exploit spurious correlations from superposition.

We emphasize that SynthSAEBench is intended to complement, not replace, LLM benchmarks. Our synthetic model cannot capture all aspects of real neural network representations (see Appendix~\ref{apx:limitations} for limitations). However, it offers capabilities that LLM benchmarks fundamentally cannot: ground-truth features, controlled ablations, low-noise metrics, and fast iteration. We envision researchers using SynthSAEBench to rapidly prototype and diagnose SAE architectures, then validating promising approaches on LLM benchmarks like SAEBench.

In future work, we hope to extend our framework to Minkowski representations~\citep{fel2025into} by introducing a ``soft'' variant of mutual exclusion via softmax, and to feature manifolds, though evaluating SAEs on manifold-structured data is challenging. More broadly, building synthetic models under different representation hypotheses and comparing SAE training dynamics to those observed in LLMs could help test the validity of those hypotheses.

\section*{Acknowledgements}
David Chanin was supported thanks to EPSRC EP/S021566/1 and the Machine Learning Alignment and Theory Scholars (MATS) program. We are grateful to Lovkush Agarwal, Fred Bruford, and Tasos Spiliotopoulos for feedback during the project.

\bibliography{references}
\bibliographystyle{icml2026}

\appendix

\section{Limitations}
\label{apx:limitations}

Synthetic data cannot capture all aspects of real neural network representations. Our generative model assumes linear feature directions, which may not hold for all concepts \citep{engels2025not}. The correlation and hierarchy structures, while configurable, are simplified approximations of the complex dependencies in real data. Most importantly, there may be ``unknown unknowns'', phenomena in real networks that we have not thought to model. Due to our lack of true ground-truth knowledge of features in LLMs, we do not know how to set hyperparameters like number of features, correlation levels, hierarchy degree, and superposition level. We also do not attempt to model complex feature geometry aside from superposition noise. Synthetic benchmarks should complement, not replace, evaluation on real models.

\section{Compensating base probabilities for hierarchy}
\label{apx:compensate}

When hierarchy constraints are applied, the effective firing probability of child features is reduced because children can only fire when their parent is active. Similarly, mutual exclusion further reduces effective probabilities since only one sibling can remain active. We implement optional probability compensation to correct for these reductions, ensuring that the \emph{effective} firing rate of each feature matches its specified base probability.

\paragraph{Hierarchy correction.} Consider a feature $i$ with base firing probability $p_i$ whose parent has base probability $p_{\text{parent}}$. Without compensation, the effective probability of feature $i$ firing is approximately $p_i \cdot p_{\text{parent}}$, since the child can only fire when the parent fires. To compensate, we scale up the sampling probability by a correction factor:
\begin{equation}
    \gamma_i^{\text{hier}} = \frac{1}{p_{\text{parent}}}.
\end{equation}
After sampling with corrected probability $\tilde{p}_i = \min(1, p_i \cdot \gamma_i^{\text{hier}})$ and applying hierarchy constraints, the effective firing rate is approximately $p_i$. The clamp at 1 is active only when $p_i > p_{\text{parent}}$; in our setup this never occurs, because features are assigned to tree nodes in breadth-first order while base probabilities follow a non-increasing Zipfian distribution, so a child always has $p_i \le p_{\text{parent}}$.

For deeper hierarchies, this correction is applied recursively. A feature at depth $d$ with ancestors having probabilities $p_1, p_2, \ldots, p_{d-1}$ would naively have effective probability $p_i \cdot \prod_{k=1}^{d-1} p_k$. However, since each ancestor also receives its own correction, the child only needs to correct for its immediate parent's base probability.

\paragraph{Mutual exclusion correction.} When a parent node has mutually exclusive children, at most one child can remain active per sample. This introduces additional probability reduction beyond hierarchy constraints.

Consider an ME group with children having base probabilities $p_1, p_2, \ldots, p_m$ under a parent with probability $p_P$. After hierarchy correction, conditioned on the parent firing, child $j$ fires with probability approximately $p_j / p_P$. When multiple children fire, one is randomly selected as the winner.

For child $i$, the expected number of competing siblings (given parent fires) is:
\begin{equation}
    \mathbb{E}[\text{competitors}] = \sum_{j \neq i} \frac{p_j}{p_P}.
\end{equation}

To first order, the probability of child $i$ being deactivated by ME is proportional to this expected competitor count. We apply a multiplicative ME correction:
\begin{equation}
    \gamma_i^{\text{ME}} = 1 + \sum_{j \neq i} \frac{p_j}{p_P}.
\end{equation}

The total correction factor for a feature in an ME group is $\gamma_i = \gamma_i^{\text{hier}} \cdot \gamma_i^{\text{ME}}$.

\paragraph{Limitations.} This compensation is approximate and assumes independence between sibling firings. In practice, correlations between features (introduced via the correlation matrix $\boldsymbol{\Sigma}$) can cause deviations from the target probabilities. The correction also does not account for higher-order effects when multiple levels of hierarchy and ME interact. Nevertheless, we find empirically that compensation substantially improves the match between specified and effective firing probabilities.

\section{Parent-scaled child magnitudes}
\label{apx:parent-scaling}

By default, hierarchy constraints apply binary gating: a child feature retains its sampled magnitude when its parent is active, and is zeroed out otherwise. We additionally support \emph{parent-scaled magnitudes}, where the child's activation is modulated by the parent's normalized activation strength. Formally, for a child feature with coefficient $c_{\text{child}}$ whose parent has coefficient $c_{\text{parent}}$ and mean firing magnitude $\bar{\mu}_{\text{parent}}$:
\begin{equation}
    c_{\text{child}} \leftarrow c_{\text{child}} \cdot \frac{c_{\text{parent}}}{\bar{\mu}_{\text{parent}}},
\end{equation}
where $\bar{\mu}_{\text{parent}}$ is the precomputed mean magnitude of the parent feature. When the parent is inactive ($c_{\text{parent}} = 0$), the child is zeroed out as in the standard case. Dividing by $\bar{\mu}_{\text{parent}}$ normalizes the scaling so that the child's expected magnitude is preserved on average.

This models the intuition that more specific concepts should be modulated by the intensity of their parent concept. For instance, features associated with specific dog breeds should fire more strongly when the ``dog'' feature is strongly active. Without this scaling, the child magnitude is independent of how strongly the parent fires, which may be unrealistic for many natural concepts.

Parent-scaled magnitudes can be enabled independently per parent node, and compose with both mutual exclusion and the probability compensation described in Appendix~\ref{apx:compensate}. The scaling is applied after binary gating and mutual exclusion resolution, so the cascading order remains: (1) parent gating, (2) mutual exclusion, (3) magnitude scaling.

\section{Low-rank correlation matrix generation}
\label{apx:correlation}

For scalability with large numbers of features, we use a low-rank representation of the correlation matrix rather than storing the full $N \times N$ matrix. This reduces storage from $O(N^2)$ to $O(Nr)$ where $r$ is the rank.

The correlation structure is represented as:
\begin{equation}
    \boldsymbol{\Sigma} = \mathbf{F}\mathbf{F}^\top + \text{diag}(\boldsymbol{\delta})
\end{equation}
where $\mathbf{F} \in \mathbb{R}^{N \times r}$ is a factor matrix and $\boldsymbol{\delta} \in \mathbb{R}^N$ contains diagonal residual variances.

\paragraph{Generation procedure.} Given rank $r$ and correlation scale $s$, we generate the factor matrix by sampling from a scaled normal distribution:
\begin{equation}
    F_{ij} \sim s \cdot \mathcal{N}(0, 1)
\end{equation}
The diagonal term is then computed to ensure unit diagonal in the implied correlation matrix:
\begin{equation}
    \delta_i = 1 - \sum_{j=1}^{r} F_{ij}^2
\end{equation}

\paragraph{Numerical stability.} If any diagonal term $\delta_i$ falls below a minimum threshold (we use 0.01), the entire factor matrix is scaled down to ensure all diagonal terms remain valid. Specifically, we compute a scale factor:
\begin{equation}
    \text{scale} = \sqrt{\frac{1 - \delta_{\min}}{\max_i \sum_j F_{ij}^2}}
\end{equation}
and apply $\mathbf{F} \leftarrow \text{scale} \cdot \mathbf{F}$, then recompute $\boldsymbol{\delta}$.

\paragraph{Efficient sampling.} Sampling from this low-rank structure requires only $O(Nr)$ computation per batch:
\begin{equation}
    \mathbf{g} = \mathbf{F}\boldsymbol{\epsilon} + \sqrt{\boldsymbol{\delta}} \odot \boldsymbol{\eta}, \quad \boldsymbol{\epsilon} \sim \mathcal{N}(\mathbf{0}, \mathbf{I}_r), \quad \boldsymbol{\eta} \sim \mathcal{N}(\mathbf{0}, \mathbf{I}_N)
\end{equation}
where $\odot$ denotes elementwise multiplication. The resulting $\mathbf{g}$ has the desired correlation structure $\mathbb{E}[\mathbf{g}\mathbf{g}^\top] = \boldsymbol{\Sigma}$.

\section{Configurable firing distributions}
\label{apx:firing-dists}

Both per-feature firing probabilities $p_i$ and per-feature magnitude parameters $(\mu_i, \sigma_i)$ admit several configurable distributions in our framework.

\paragraph{Firing probabilities.} The distribution of $p_i$ across features significantly impacts SAE training dynamics: very rare features are difficult to learn, while very frequent ones dominate reconstruction loss. We use a Zipfian distribution as the default, motivated by the observation that concept frequencies in natural data follow power laws~\citep{piantadosi2014zipf,ayonrinde2024adaptive,michaud2023quantization}: $p_i \propto i^{-\alpha}$, scaled to lie in $[p_{\min}, p_{\max}]$. We also support linear decay ($p_i$ interpolates linearly from $p_{\max}$ to $p_{\min}$), uniform random ($p_i \sim \text{Uniform}(p_{\min}, p_{\max})$), and constant ($p_i = p$). Arbitrary distributions can be plugged in.

\paragraph{Firing magnitudes.} When a feature fires, its coefficient is sampled from a rectified Gaussian: $c_i = \text{ReLU}(\mu_i + \sigma_i \epsilon_i)$, $\epsilon_i \sim \mathcal{N}(0,1)$. The per-feature mean $\mu_i$ and standard deviation $\sigma_i$ can each take constant, linear-interpolated, exponential-interpolated, or folded-normal forms (e.g., $\sigma_i \sim |\mathcal{N}(\mu_\sigma, \sigma_\sigma^2)|$); the folded normal creates heterogeneous magnitude variability. Arbitrary distributions over $\mu_i$ and $\sigma_i$ are supported.

\section{SAE training procedures}
\label{apx:training}

We train all SAEs using the Adam optimizer~\citep{kingma2014adam} with learning rate $3 \times 10^{-4}$, $\beta_1 = 0.9$, $\beta_2 = 0.999$, and batch size 1024. The learning rate decays linearly to zero over the final third of training. All SAEs have width 4096 and are trained on 200M samples from SynthSAEBench. Below we describe architecture-specific training details.

\subsection{BatchTopK SAEs}
\label{apx:batchtopk}

BatchTopK SAEs~\citep{bussmann2024batchtopk} use a soft top-$k$ selection that allows the number of active features to vary per sample while maintaining a target average L0 across the batch. The target L0 is set directly via the $k$ parameter, so no autotuning is required.

To prevent dead latents, we use the TopK auxiliary loss from \citet{gao2024scaling}. This auxiliary loss has dead latents reconstruct the residual error from live latents, providing gradient signal to features that would otherwise receive none. Following the heuristic from \citet{gao2024scaling}, we set $k_{\text{aux}} = D/2$ where $D$ is the input dimension, and scale the loss by $\min(\text{num\_dead} / k_{\text{aux}}, 1)$ to reduce its magnitude when few latents are dead.

\subsection{JumpReLU SAEs}
\label{apx:jumprelu}

Our JumpReLU training procedure follows the original procedure in \citet{rajamanoharan2024jumping} matching the procedure used by SAEBench, except we tune some hyperparameters to perform better in the synthetic setting. We use initial JumpReLU threshold of 1.0, bandwidth of 1.0, and use a latent norm of 0.5 at initialization. We find that with these settings, JumpReLU SAEs perform well on SynthSAEBench and we do not see many dead latents.

For sparsity control, we use an initial penalty of 1.0 with no warm-up and immediately adjust using an L0 coefficient autotuner (Appendix~\ref{apx:autotune}) to hit a target L0.

\subsection{Standard L1 SAEs}
\label{apx:l1}

For standard L1 SAEs, we follow the procedure from \citet{olah2024april}, training with a weighted combination of reconstruction MSE and an L1 penalty on feature activations. The L1 coefficient is warmed up linearly over the first third of training, after which we use the L1 coefficient autotuner (Appendix~\ref{apx:autotune}) to achieve target L0 values. No auxiliary loss is used.

\subsection{Matryoshka SAEs}
\label{apx:matryoshka}

We train Matryoshka SAEs~\citep{bussmann2025learning} using the BatchTopK activation function with nested prefixes. We use prefix sizes $\mathcal{M} = \{128, 512, 2048, 4096\}$. Each prefix is trained to reconstruct the input independently.

While the original Matryoshka SAEs work uses the standard TopK SAE auxiliary loss described in Appendix~\ref{apx:batchtopk}, we find that using a Matryoshka-optimized auxiliary loss results in better performance, especially at low L0.

\paragraph{Matryoshka auxiliary loss.} The standard TopK auxiliary loss~\citep{gao2024scaling} trains dead latents to reconstruct the residual error of the full SAE. However, in a Matryoshka SAE, dead latents at early prefixes (e.g., the first 128 latents) face a very different reconstruction residual than dead latents at later prefixes. Training all dead latents against the full SAE's residual provides a poor learning signal for latents in smaller prefixes, since those latents need to help reconstruct a much larger residual.

Our Matryoshka auxiliary loss instead computes a separate auxiliary loss for each matryoshka prefix. For prefix $m \in \mathcal{M}$, let $\hat{a}_m$ be the reconstruction using the first $m$ latents, and let the dead latents within the range $[m_{\text{prev}}, m)$ be denoted $\mathcal{D}_m$. The auxiliary loss for prefix $m$ is:
\begin{equation}
    \mathcal{L}_{\aux,m} = s_m \cdot \left\| W_{\dec,\mathcal{D}_m} f_{\aux,m} - (a - \hat{a}_m)_{\text{detach}} \right\|_2^2,
\end{equation}
where $f_{\aux,m}$ are the top-$\min(k_{\aux}, |\mathcal{D}_m|)$ activations among dead latents in $[m_{\text{prev}}, m)$, using only the corresponding portion of the encoder pre-activations. The scale factor $s_m = \min(|\mathcal{D}_m| / k_{\aux}, 1)$ reduces the loss magnitude when few latents in that prefix are dead. The residual $(a - \hat{a}_m)$ is detached to prevent the auxiliary loss from affecting the main reconstruction pathway.

This per-prefix formulation ensures that dead latents in early prefixes receive gradient signal appropriate to their level's reconstruction error, rather than the much smaller residual of the full SAE.

We explore dead latents further in Appendix~\ref{apx:dead-latents}.

\section{Autotuning sparsity coefficients}
\label{apx:autotune}

For SAE architectures that use a sparsity-inducing loss (standard L1 SAEs and JumpReLU SAEs), the sparsity coefficient $\lambda$ controls the trade-off between reconstruction and sparsity. However, the relationship between $\lambda$ and the resulting L0 sparsity is nonlinear and model-dependent, making it difficult to train SAEs at a specific target L0 for fair comparison. We implement an autotuning controller that dynamically adjusts a multiplier $m$ on the sparsity coefficient during training to achieve a target L0.

\paragraph{Controller design.} We use a rate-dampened integral controller~\citep{aastrom2021feedback} that adjusts the effective sparsity coefficient $\lambda_{\text{eff}} = \lambda \cdot m$ based on the deviation from target L0. The controller maintains:

\begin{itemize}
    \item A smoothed L0 estimate $\bar{\ell}_t$ using exponential moving average (EMA) with smoothing factor $\alpha$
    \item A smoothed rate of L0 change $\dot{\ell}_t$ (the derivative estimate)
\end{itemize}

At each training step, given batch L0 measurement $\ell_t$:
\begin{align}
    \bar{\ell}_t &= \alpha \bar{\ell}_{t-1} + (1 - \alpha) \ell_t \\
    \dot{\ell}_t &= \alpha_r \dot{\ell}_{t-1} + (1 - \alpha_r) (\bar{\ell}_t - \bar{\ell}_{t-1})
\end{align}
where $\alpha = 0.99$ and $\alpha_r = 0.95$ are the smoothing factors for position and rate respectively.

\paragraph{Gain scheduling.} Gain scheduling~\citep{rugh2000research} adapts controller parameters based on operating conditions. The key insight is that when the system is converging toward the target (error decreasing), we should reduce the controller gain to prevent overshoot. We detect convergence when the error and rate have opposite signs:
\begin{equation}
    \text{converging} = (\bar{\ell}_t - \ell^*) \cdot \dot{\ell}_t < 0
\end{equation}
where $\ell^*$ is the target L0. When converging, we reduce the gain by a factor $\gamma_c = 0.01$.

\paragraph{Bounded adjustment.} To ensure stable behavior, we use a tanh nonlinearity to bound the adjustment magnitude:
\begin{equation}
    \Delta = K_i \cdot g \cdot \tanh\left(\left|\frac{\bar{\ell}_t - \ell^*}{\ell^*}\right| \cdot s\right)
\end{equation}
where $K_i = 3 \times 10^{-4}$ is the integral gain, $g \in \{\gamma_c, 1\}$ is the scheduled gain, and $s = 10$ is the gain scale. The multiplier is then updated multiplicatively:
\begin{equation}
    m_{t+1} = \begin{cases}
        m_t \cdot (1 + \Delta) & \text{if } \bar{\ell}_t > \ell^* \\
        m_t \cdot (1 - \Delta) & \text{if } \bar{\ell}_t < \ell^*
    \end{cases}
\end{equation}
The multiplier is clamped to $[0.01, 100]$ to prevent extreme values.

\paragraph{Integration with SAE training.} For standard L1 SAEs, the autotuner modulates the L1 coefficient: $\lambda_1^{\text{eff}} = \lambda_1 \cdot m$. For JumpReLU SAEs, it modulates the L0 penalty coefficient similarly. The autotuner state is updated after each training batch, and the new multiplier is applied to the next batch's loss computation.

This approach allows us to train SAEs at precisely matched L0 values across different architectures, enabling fair comparison of their reconstruction and feature recovery quality.

\section{Dead latents in SynthSAEBench}
\label{apx:dead-latents}

Dead latents are a problem in SAE training in general, with SynthSAEBench being no exception. We evaluate dead latents over 1M samples for the experiments in this section. 

\subsection{Dead latents in BatchTopK and Matryoshka SAEs}

\begin{figure}[ht]
    \centering
    \includegraphics[width=0.5\columnwidth]{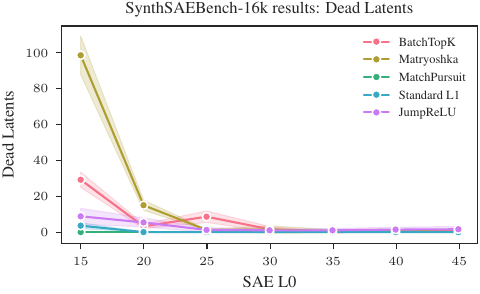}
    \caption{Dead latents vs L0 for SynthSAEBench SAEs. Shaded area is 1 stdev with 5 random seeds.}
    \label{fig:dead-latents-synth-16k}
\end{figure}

BatchTopK and especially Matryoshka BatchTopK SAEs seem to struggle with dead latents at low L0 in SynthSAEBench, even with the aux loss reviving dead latents. We can see this in Figure~\ref{fig:dead-latents-synth-16k}, where dead latents increase at L0=15 for BatchTopK and especially Matryoshka SAEs. We find that these latents are actually not entirely ``dead'', as they will still fire occasionally, but far less frequently than our dead latent window. This suggests that SAEs are settling into poor local minima.

\begin{figure*}[ht]
    \centering
    \includegraphics[width=\linewidth]{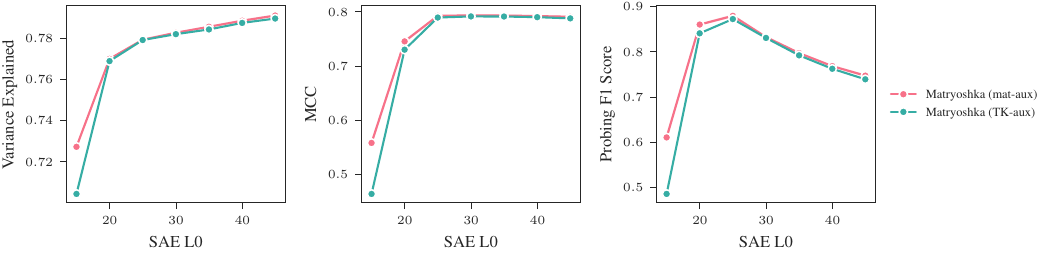}
    \caption{Comparing Matryoshka SAEs trained with standard TopK auxiliary loss (TK-aux) and a Matryoshka-optimized TopK auxiliary loss (mat-aux). The matryoshka-optimized loss results in better SAEs, but the difference is especially pronounced at low L0.}
    \label{fig:mat-aux-loss}
\end{figure*}

\begin{figure}[ht]
    \centering
    \includegraphics[width=0.5\linewidth]{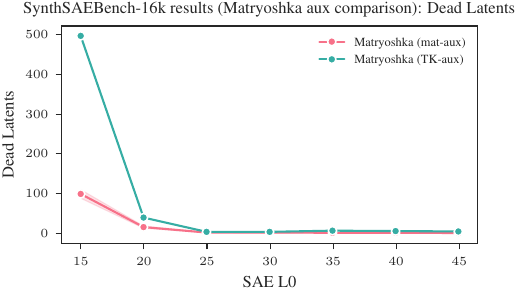}
    \caption{Dead latents for Matryoshka SAEs trained with standard TopK auxiliary loss (TK-aux) and a Matryoshka-optimized TopK auxiliary loss (mat-aux). The matryoshka-optimized loss results in fewer dead latents, especially pronounced at low L0.}
    \label{fig:mat-aux-loss-dead-latents}
\end{figure}

We also find that using the Matryoshka-optimized variant of the auxiliary loss described in Appendix~\ref{apx:matryoshka} results in a better Matryoshka SAE with fewer dead latents, especially at low L0. We show SAE quality results comparing these auxiliary losses for Matryoshka SAEs in Figure~\ref{fig:mat-aux-loss}, and show dead latents in Figure~\ref{fig:mat-aux-loss-dead-latents}.

\subsection{Dead latents in Matching Pursuit and Standard L1 SAEs.}

Matching Pursuit (MP) SAEs seem to never have any dead latents in SynthSAEBench, no matter the setting. This is impressive and a clear benefit of this architecture. Standard L1 SAEs also did not have any problems with dead latents in our experiments as long as the L1 coefficient is linearly warmed up as suggested by \citet{olah2024april}.

\section{SynthSAEBench sample generation performance}
\label{apx:perf}

\begin{figure}
    \centering
    \includegraphics[width=0.5\linewidth]{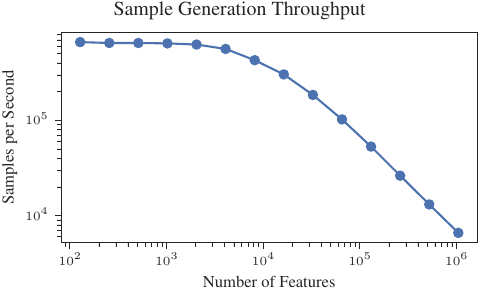}
    \caption{Synthetic model throughput by number of features.}
    \label{fig:throughput}
\end{figure}

We next investigate the performance characteristics of the synthetic data generation process as a function of number of features in the synthetic model. We use our base SynthSAEBench model, but vary the number of features in the model from $2^7$ (128) to $2^{20}$ (1M). We keep the same 3-level mutually-exclusive hierarchy scaled relative to the size of the data model. We then sample 100 batches of size 1024 on an Nvidia H100 GPU and benchmark the sample throughput of the model. Results are shown in Figure~\ref{fig:throughput}.

Feature generation ranges from 600K samples / sec (2.5 min for 100M samples) for models with under 1000 features to 7K samples / sec (4hr for 100M samples) for a model with 1M features. We decided on 16k as a good compromise between these two extremes. At 16k features, the model generates at 300K samples / sec, or 5 min for 100M samples.

\section{Noise in SAEBench metrics}
\label{apx:noise-in-saebench-metrics}

\begin{figure*}[ht]
    \centering
    \includegraphics[width=\linewidth]{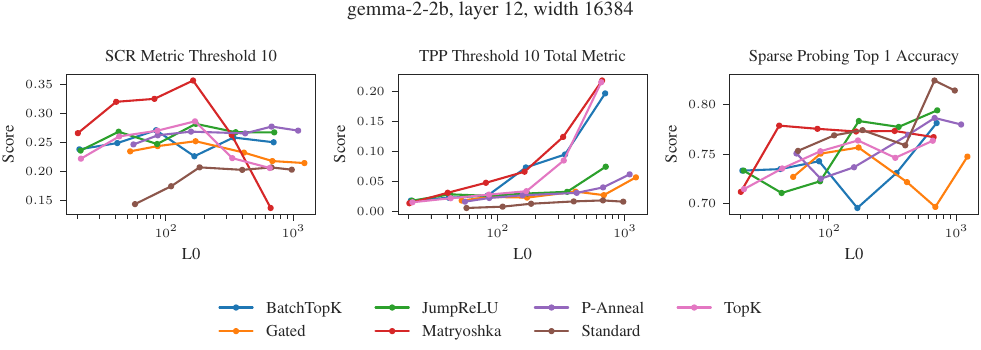}
    \caption{SAEBench SCR, TPP, and Sparse Probing metrics for Gemma-2-2b layer 12 width 16k SAEs.}
    \label{fig:noise-in-saebench-metrics}
\end{figure*}

One motivation for this work is that existing SAE benchmarks like SAEBench, while extremely important and indispensable for SAE architecture development, tend to be noisy. For instance, we show SAEBench Spurious Correlation Removal (SCR), Targeted Probe Perturbation (TPP), and Sparse Probing metrics from the SAEBench paper for Gemma-2-2b layer 12 width 16k SAEs in Figure~\ref{fig:noise-in-saebench-metrics}.

While there are clear trends in some metrics, overall there is still a lot of random noise that makes it difficult to make fine-grained judgements about SAE architecture differences. Cutting through this noise requires running many seeds of SAEs, which is often infeasibly expensive for LLM SAEs.

\section{Exploring Superposition}
\label{apx:exploring-superposition}

\begin{figure}[ht]
    \centering
    \includegraphics[width=0.85\linewidth]{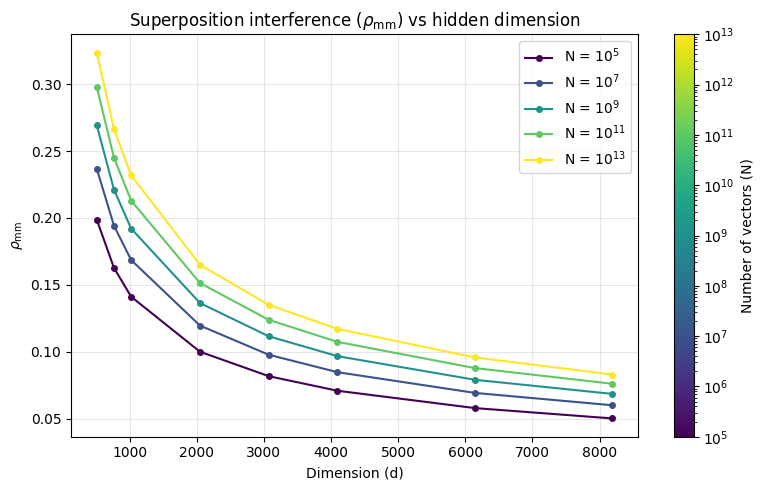}
    \caption{Mean max absolute cosine similarity $\rho_{\text{mm}}$ as a function of hidden dimension $D$, computed analytically from Eq.~\eqref{eq:rho-mm-cdf} for random unit-norm feature directions, with the number of features $N$ ranging from $10^5$ to $10^{13}$.}
    \label{fig:exploring-superposition}
\end{figure}

How much pairwise feature interference does a given hidden dimension impose, and how does it scale with the number of features? Below we give a closed-form characterisation of $\rho_{\text{mm}}$ for random unit-norm directions, which is essentially indistinguishable from the optimum in the regimes relevant to neural networks.

\paragraph{Distribution of pairwise cosine similarity.} For two independent unit-norm vectors $u, v \in \mathbb{R}^D$ drawn uniformly from the sphere, the squared cosine similarity $\langle u, v\rangle^2$ follows a $\mathrm{Beta}(1/2, (D-1)/2)$ distribution. Letting $F_D$ denote its CDF, the maximum absolute cosine similarity of one direction against the other $N-1$ directions has CDF
\begin{equation}
\label{eq:rho-mm-cdf}
P\!\left(\max_{j \ne i}\,|\langle u_i, u_j\rangle| \le x\right) \;=\; \bigl[F_D(x^2)\bigr]^{N-1},
\end{equation}
and $\mathbb{E}[\rho_{\text{mm}}]$ follows by direct integration, with no Monte Carlo simulation required. Spherical packing bounds further show that, in the regime $N \ll 2^D$ that covers all practical neural network widths, optimally-placed directions achieve essentially the same $\rho_{\text{mm}}$ as random ones, so the analysis applies whether or not features are explicitly orthogonalised.

\paragraph{Asymptotic scaling.} Taylor expanding Eq.~\eqref{eq:rho-mm-cdf} around $x = 0$ yields the simple approximation
\begin{equation}
\rho_{\text{mm}} \;\approx\; \sqrt{\frac{\ln N}{D}}.
\end{equation}
Two consequences follow. First, interference shrinks as $O(1/\sqrt{D})$ in the hidden dimension. Second, the dependence on the number of features is only logarithmic: scaling $N$ by $100\times$ adds only $\sqrt{\ln 100} \approx 2.1$ to the numerator, an effect easily absorbed by a modest increase in $D$. Hidden dimension dominates feature count in determining how much superposition is feasible.

\paragraph{Implications for SynthSAEBench.} Figure~\ref{fig:exploring-superposition} plots $\rho_{\text{mm}}$ from Eq.~\eqref{eq:rho-mm-cdf} against $D$ for $N$ ranging from $10^5$ to $10^{13}$. Even at $N = 10^{13}$, $\rho_{\text{mm}}$ falls below $0.1$ once $D \geq 6000$. Conversely, SynthSAEBench ($N = 16{,}384$, $D = 768$) yields $\rho_{\text{mm}} \approx 0.15$, which is substantially more pairwise interference than would arise for trillions of features at a modern LLM hidden dimension. This is roughly equivalent to the amount of superposition we would expect in 1 Billion features in 2k hidden dim, roughly the size of Gemma-2-2b, which we feel is a reasonable baseline.

\section{Logistic regression probes on SynthSAEBench}
\label{apx:probes}

In Section~\ref{sec:prec-recall}, we note that the best SAE architecture (Matryoshka BatchTopK) achieves a peak probing F1 of approximately 0.88, consistent with LLM SAE findings that SAEs underperform supervised probes~\citep{kantamneni2025sparse}. To quantify this gap on SynthSAEBench, we train logistic regression probes directly on hidden activations to classify ground-truth feature firings.

\paragraph{Setup.} We sample 2M activations from SynthSAEBench and evaluate the first 4,096 features, which are the highest-frequency features due to the Zipfian ordering of firing probabilities. We train one linear probe per feature simultaneously using a batched logistic regression model (a single weight matrix $W \in \mathbb{R}^{F \times D}$ and bias $b \in \mathbb{R}^F$, where $F$ is the number of probed features). Training uses Adam with learning rate $3 \times 10^{-3}$, cosine annealing over 10,000 steps, batch size 4,096, $L_2$ weight decay of $10^{-3}$, and class-imbalance-corrected binary cross-entropy loss. We use an 80/20 train/test split, and tune a per-feature classification threshold on a 200K-sample validation subset drawn from the training set by sweeping 499 thresholds and selecting the one that maximizes F1 per feature.

\paragraph{Results.} Results are shown in Table~\ref{tab:probe-results} in the main text. The probes achieve a mean F1 of 0.974 and mean AUC of 0.9999, substantially outperforming the best SAE probing F1 of $\approx$0.88. This confirms that the gap between SAE probing and supervised probing observed in LLMs~\citep{kantamneni2025sparse} is reproduced in SynthSAEBench, and that this gap is not an artifact of LLM evaluation noise but reflects a genuine limitation of current SAE architectures.

\section{Extended results}
\label{apx:extended}

We include extended results for the experiments in the paper.

\subsection{SynthSAEBench extended results}
\label{apx:synthsaebench-extended}

\begin{figure*}[ht]
    \centering
    \includegraphics[width=\linewidth]{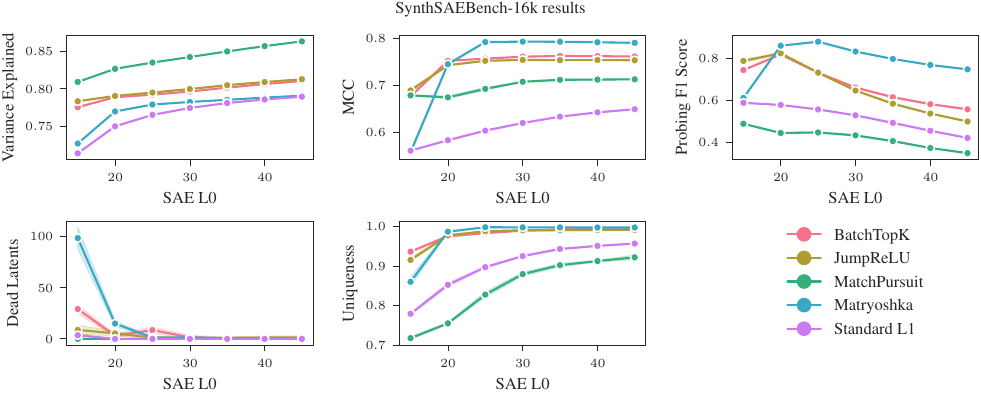}
    \caption{Full results for SynthSAEBench. Shaded area represents 1 stdev with 5 seeds per SAE.}
    \label{fig:vary-l0-extended}
\end{figure*}

Full results for SynthSAEBench are shown in Figure~\ref{fig:vary-l0-extended}. 

\subsection{Ablation: feature correlation strength}

\begin{figure*}[ht]
    \centering
    \includegraphics[width=\linewidth]{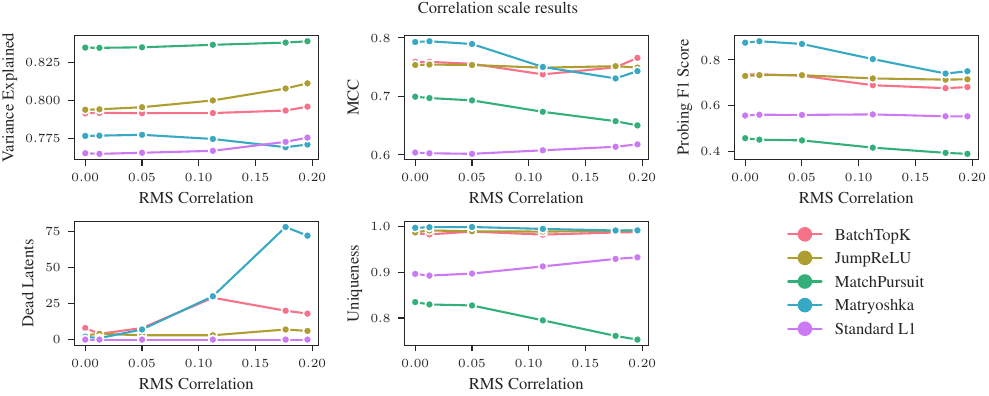}
    \caption{Results varying correlation strength, while keeping the remaining model hyperparameters set at default values for SynthSAEBench-16k.}
    \label{fig:vary-correlation-extended}
\end{figure*}

Next, we explore the effect of varying the strength of the random correlations between features. We vary the correlation scale used to generate the random correlation matrix from 0 to 0.25, and train SAEs with L0=25. Results are shown in Figure~\ref{fig:vary-correlation-extended}.

The effect of varying correlation strength are less dramatic than varying superposition, but still interesting. We see that variance explained increases for all SAEs as correlation increases except for Matryoshka SAEs, but this appears to be due to dead latents rather than a fundamental architectural issue. This is consistent with previous work showing that SAEs are able to exploit feature correlation to increase reconstruction by mixing correlated features~\citep{chanin2025feature,chanin2025sparse}. We see slight decreases in probing F1 score as correlation increases, but mixed results on MCC that are harder to judge.

Likely the effect of feature correlation in SynthSAEBench is overshadowed by the effect of superposition noise and hierarchy (itself an even more extreme form of correlation).

\subsection{Ablation: feature correlation rank}

\begin{figure*}[ht]
    \centering
    \includegraphics[width=\linewidth]{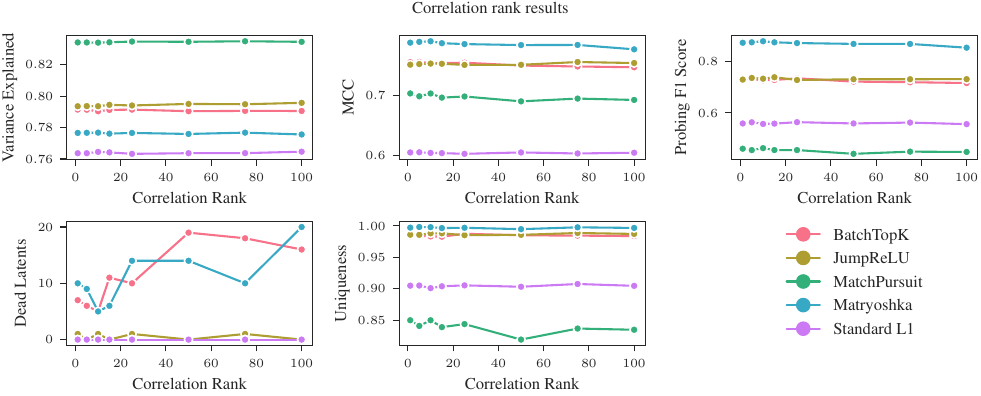}
    \caption{Results varying correlation rank, with correlation strength 0.075, and keeping the remaining model hyperparameters set at default values for SynthSAEBench-16k.}
    \label{fig:vary-correlation-rank}
\end{figure*}

Next, we explore the effect of varying the rank of the low-rank correlations between features. We vary the rank from 1 to 100, and train SAEs with L0=25. We use correlation strength 0.075 throughout. Results are shown in Figure~\ref{fig:vary-correlation-rank}.

The effect of varying correlation rank is effectively negligible. This is not surprising, since the correlation rank just changes the types of correlation patterns that are possible, but does not change the underlying strength or quantity of correlations.

\subsection{Ablation: feature firing probability}

\begin{figure*}[ht]
    \centering
    \includegraphics[width=0.8\textwidth]{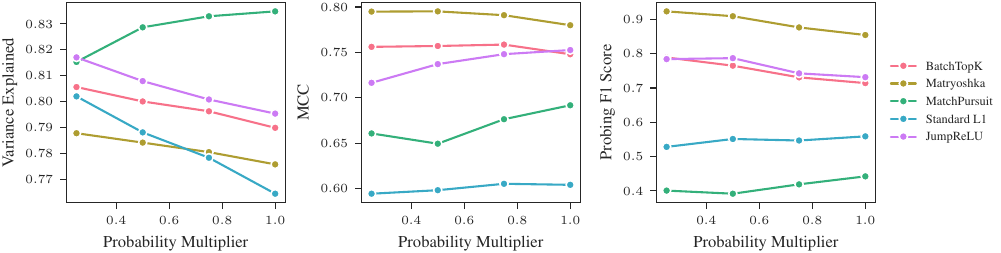}
    \caption{Ablation study varying the base firing probabilities of the SynthSAEBench model. We scale the L0 of the SAEs to match the scaled firing probabilities based on the default L0=25 (L0=12.5 when probability multiplier is 0.5). We see broadly the same pattern as in the main experiments, with Matryoshka being best at quality metrics, and Matching Pursuit having the best reconstruction. Interestingly, Matching Pursuit SAEs increase variance explained with higher L0 while other SAEs do not.}
    \label{fig:vary-firing-prob}
\end{figure*}

We perform an ablation study where we decrease the base firing probabilities of all features by a multiplier, effectively decreasing the L0 of the model. We evaluate each SAE at L0$=25 * p$ where $p$ is the firing multiplier. Results are shown in Figure~\ref{fig:vary-firing-prob}. We see broadly the same pattern as in the main experiments, with Matryoshka being best at quality metrics, and Matching Pursuit having the best reconstruction. Interestingly, Matching Pursuit SAEs increase variance explained with higher L0 while other SAEs do not.

\subsection{Ablation: feature firing stdev}

\begin{figure*}[ht]
    \centering
    \includegraphics[width=0.8\textwidth]{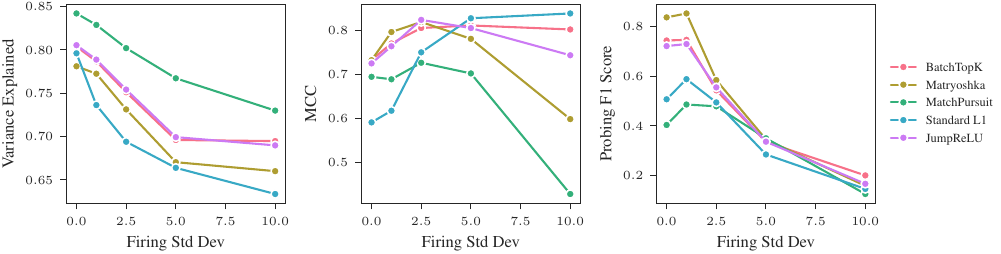}
    \caption{Ablation study varying the feature firing magnitude stdev of the SynthSAEBench model. We see broadly the same pattern as in the main experiments, with Matryoshka being best at quality metrics, and Matching Pursuit having the best reconstruction. Interestingly, we see a dichotomy between MCC and F1 score at high firing stdev, where high stdev improves MCC but reduces F1 score for all SAEs. This makes sense as high firing magnitude variance makes feature directions more prominent for the SAE, but means there are more low-magnitude samples that are difficult for the SAE to disambiguate from superposition noise.}
    \label{fig:vary-stdev}
\end{figure*}

We perform an ablation study where we set all features to have the same firing magnitude stdev, and then vary the stdev from 0.5 to 10.0. Each SAE is trained at L0=25. Results are shown in Figure~\ref{fig:vary-stdev}. At high stdev, F1 score and variance explained drops for every SAE. This makes sense since it is hard to distinguish a low-magnitude firing from superposition noise when there is such a wide range of potential firing magnitudes for each feature. Interestingly, however, standard L1 SAEs achieve the best MCC of all SAEs at high firing magnitude stdev. We are not sure why this is, but it could be worth studying further to try to understand why a simple L1 penalty does so well in this regime compared to the other SAE architectures.

\subsection{Ablation: deeper hierarchy}

\begin{figure*}[ht]
    \centering
    \includegraphics[width=0.8\textwidth]{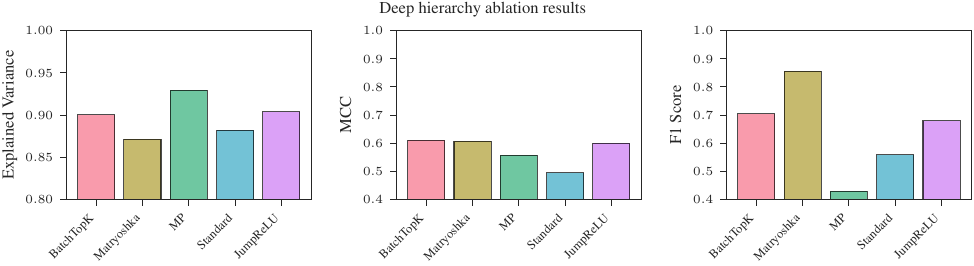}
    \caption{Ablation study using a much deeper hierarchy (128 root nodes, 2 children per node, max depth of 6). All SAEs are trained at L0=25. We see the same general pattern as in SynthSAEBench.}
    \label{fig:deep-hierarchy}
\end{figure*}

We perform an ablation study using a much deeper hierarchy (128 root nodes, 2 children per node, max depth of 6). All SAEs are trained at L0=25. Results are shown in Figure~\ref{fig:deep-hierarchy}. We see the same general pattern as in SynthSAEBench, where the Matryoshka SAE is best by quality metrics (MCC and F1), despite poor reconstruction (variance explained). The MP-SAE achieves best reconstruction, but scores poorly on quality metrics. JumpReLU and BatchTopK are in-between on both SAE quality and reconstruction, and the Standard SAE performs poorly on all metrics.

\end{document}